\documentclass{article}
\usepackage{spconf,amsmath,graphicx}
\usepackage{algorithm}
\usepackage{algorithmic}
\usepackage{booktabs}
\usepackage{caption}
\usepackage{pgfplots}
\usepackage{pgfplotstable}
\usepackage{subcaption}
\usepackage{multirow}
\usepackage{acronym}
\usepackage{tabularx}
\newacro{CNN}[CNN]{Convolutional Neural Network}
\newacro{BoCF}[BoCF]{Bag of Color Features}
\newacro{FC4}[FC$^4$]{Fully Convolutional Color Constancy}
\newacro{MCDE}[MCDE]{Monte Carlo Dropout Ensembles}

\title{Effective prevention of semantic drift as angular distance in memory-less continual deep neural networks}
\twoauthors{Khouloud Saadi
                       \thanks{This work was supported by Fatima Al-Fihri predoctoral fellowship program (https://fatimafellowship.com/)}}
                            {Independant Researcher \\
                            saadikhouloud2@gmail.com}
                            {Muhammad Taimoor Khan}
                            {Department of Computer Science, \\
                             National University of Computer and Emerging Sciences, Pakistan\\
                             taimoor.khan@nu.edu.pk}

\begin{document}
\ninept
\maketitle
\begin{abstract}
Lifelong machine learning or continual learning models attempt to learn incrementally by accumulating knowledge across a sequence of tasks. Therefore, these models learn better and faster. They are used in various intelligent systems that have to interact with humans or any dynamic environment e.g., chatbots and self-driving cars. Memory-less approach is more often used with deep neural networks that accommodates incoming information from tasks within its architecture. It allows them to perform well on all the seen tasks. These models suffer from semantic drift or the plasticity-stability dilemma. The existing models use Minkowski distance measures to decide which nodes to freeze, update or duplicate. These distance metrics do not provide better separation of nodes as they are susceptible to high dimensional sparse vectors.     
In our proposed approach, we use angular distance to evaluate the semantic drift in individual nodes that provide better separation of nodes and thus better balancing between stability and plasticity. The proposed approach outperforms state-of-the art models by maintaining higher accuracy on standard datasets. 

\end{abstract}
\section{Introduction}

Machine learning has become more and more important in today's industries. We can notice the great success of machine learning and its considerable achievements in various domains e.g., construction, robotics, and medicine~\cite{chen2018lifelong}. The current dominant machine learning approach consists of training a model in isolation. It requires to collect huge amount of data, build a model and train it on the available sample data. These trained models are deployed to facilitate end users as real-life applications~\cite{d2019episodic}. In isolated learning, the model is taken out of service and retrained each time there is new data available. For many tasks the data and computational resources are not available in sufficient supply to train complex models. Moreover, in case of multiple tasks, the isolated learning mechanism does not share the common features across those tasks and rather each time learn it from scratch.

\begin{figure}[t]
\centering
\includegraphics[width= 0.5\textwidth]{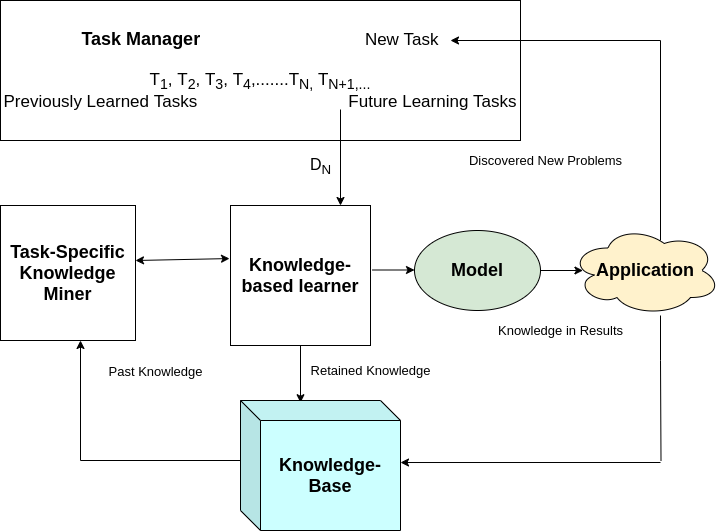}
\caption{Architecture of LML model that maintains a knowledge-base which is updated after performing each task to accumulate knowledge continuously~\cite{chen2018lifelong}.}
\label{intro_fig}
\end{figure}

In contrast, a lifelong machine learning (LML) model attempts to mimic the way we humans learn~\cite{khan2017three}. An LML based model learns incrementally by building on top of the knowledge acquired from the tasks performed in the past. The knowledge across multiple tasks accumulates to enhance learning of newer tasks without dropping its performance on the previously performed tasks. The working mechanism of a lifelong learning model is depicted in \textbf{Fig.~\ref{intro_fig}}. For the current task $T_{N+1}$, the model accumulates its underlying dataset $D_{N+1}$ and relevant knowledge from the previous $T_1$ to $T_N$ tasks. The updated model is available for the application to perform any of the $T_1$ to $T_{N+1}$ tasks. Basic features are expected to be common across multiple tasks where more overlap is expected as the number of tasks increase. 
For example, the language structures can be used in multiple tasks like question answering, sentiment analysis and summary generation. The provision of relevant knowledge by the knowledge miner ensures that the model does not relearn what is already learnt and rather improve on it~\cite{khan2016online, khan2018trends}.

Considering the importance of continual learning in diverse areas, it has variants in prominent machine learning paradigms i.e., supervised, unsupervised and reinforcement learning~\cite{chen2018lifelong}. Continual learning empowers to have reduced computational cost, and have lesser dependency on the quantity and quality of underlying data for a task. Most importantly it helps to avoid relearning the same patterns repeatedly from scratch in performing different tasks~\cite{parisi2019continual}. Transfer learning, meta learning, online learning and multi-task learning are the other paradigms that are related to continual learning~\cite{khan2017three, khan2018trends}. However, transfer learning and meta learning performs single step and unidirectional transfer of knowledge from a single or multiple source tasks to the target task. Online learning continuously learns about the changing needs of a task by updating itself with a continuous influx of training data at the cost of unlearning the previously acquired knowledge~\cite{fontenla2013online}. Multi-task learning is the most related of all learning paradigms but it expects all the tasks to be available at once in order to facilitate learning about multiple tasks in parallel~\cite{caruana1997multitask}. But the luxury of having access to all the tasks and their data is least practical in many application areas. Thus, LML or continual learning overshadows the existing learning paradigms by supporting incremental learning on a sequence of tasks. The learning is persistent and long term that allows to learn newer tasks without allowing to degrade its performance on the previously performed tasks~\cite{wu2019large}. The phenomenon of forgetting previously acquired information in order to learn new information is termed as catastrophic forgetting or semantic drift. Continual learning models attempt to counter the semantic drift and rather utilize the previous information to facilitate learning newer tasks. Its significance increases many-folds in case of tasks with limited or noisy data.    

There are two types of approaching in continual learning i.e., memory-based and memory-less. The memory-based LML models maintain an external knowledge base where knowledge from a sequence of tasks is accommodated~\cite{khan2017three}. It has modules that deal with harvesting useful knowledge, representing them in a compact form, and maintaining them for long-term use~\cite{lopez2017gradient}. These modules are activated before and after performing each task to provide relevant past knowledge and update the knowledge-base, respectively, as in Figure \ref{intro_fig}.  
The memory-less LML models are often used with deep learning, also called continual deep learning models. It revisits the architecture of the model after performing each task in order to add the new information of the current task within the model~\cite{rusu2016progressive}. Therefore, the model architecture updates and expands after each task. Sequentially trained on $N$ tasks, the model is expected to perform well on all the tasks. The memory-less approach for continual learning suits deep neural networks better as the implicit nature of feature representation within the weights of the network cannot be connected with explicit external knowledge-base. 

The memory-less continual deep learning models separate the nodes of the existing network architecture into freeze, partially regularize or duplicate categories for accommodating the current task. For the current task that is very similar to the previous tasks, majority of the nodes are frozen while others are partially regularized. Whereas, in case of high difference most of the nodes are duplicated to learn new information without unlearning the information about the previous tasks. Freezing and regularizing too many nodes restrict the model to settle for a sub-optimal performance while duplicating too many nodes limit information sharing and allows the model to grow rapidly~\cite{rusu2016progressive, kirkpatrick2017overcoming}. In the literature of continual learning this problem is known as the plasticity-stability dilemma. 
The existing research in continual deep learning use magnitude based distance measures as objective function to evaluate semantic drift or catastrophic forgetting in a node represented by a vector of its parameters~\cite{yoon2017lifelong}. In this paper we argue that measures such as euclidean and manhattan distance, as a family of Minkowski metrics, are not the optimal choice for this task. They are sensitive to scale and sparsity where all the points with high dimensional feature space become uniformly distant~\cite{aggarwal2001surprising}.  
In this research, we have used angular distance between the weight vectors of a node i.e., ${W}_1$ and ${W}_2$ to measure the semantic drift in a node as $angular-dis(W_1, W_2) = \arccos(\frac{W_1 . W_2}{||W_1||\times ||W_2||})\times (180/\pi)$. 
With known intervals angular distance is better for applying thresholds to separate nodes to freeze, partially regularize and duplicate. The proposed approach outperforms state-of-the art approaches by maintaining higher accuracy above 96\% and 97\% on standard continual deep learning datasets i.e., permuted-MNIST, permuted-Fashion-MNIST, and rotated-MNIST. 

\section{Related work} \label{related}
Continual learning systems can function in a dynamic open environment where they can discover new problems which form new tasks. these new tasks need to be learned in real-time without forgetting the old tasks~\cite{chen2018lifelong, d2019episodic}. This is called learning on the job as it attempts to achieve artificial general intelligence by optimizing across multiple tasks~\cite{khan2016online, khan2018trends}. The problem of catastrophic forgetting or semantic drift can be observed in transfer learning based models where once the weights of the model are updated according to the target task, the performance on the trained task is expected to degrade. Continual learning addresses this problem by modifying the model architecture such that it learns newer tasks but also retains its learning of the previous tasks as well~\cite{yoon2017lifelong}. However, allowing the model to learn more results in losing the previously learnt tasks and vice versa. This problem is known as the plasticity-stability dilemma in the literature of continual deep learning models~\cite{d2019episodic}.

Memory-based continual learning models maintain an extra memory unit to store refined information across all previously performed tasks. In case of gradient of episodic memory which is based on episodic memory, several examples are stored from past tasks and used on minimizing negative backward transfer while learning new tasks~\cite{lopez2017gradient}. Incremental classifier and representation learning use a combination of two losses i.e., the ordinary classification loss of the current task and distillation loss which will encourage the learner to remember the previous tasks in a sequence~\cite{rebuffi2017icarl}. It uses a subset of exemplars stored from previous tasks to perform a nearest class mean classifier. The other type i.e., the memory-less continual learning models accommodate the incoming information within the model architecture. It follows two working mechanisms. The first is based on regularization e.g., elastic weight consolidation~\cite{kirkpatrick2017overcoming}. It identifies the weights that are specific to the previous tasks and penalizes their update in order to retain high performance on the previous tasks. Therefore, the model maintains high stability. In case of learning without forgetting, all the parameters i.e., both generic and task specific are updated within a constraint in such a way that the model learns newer tasks without compromising its performance on the previous tasks~\cite{li2017learning}. The second type involves dynamic architectures such as progressive neural networks that grows progressively for each task by adding a certain number of nodes to accommodate the new features while freezing the nodes of past tasks for better stability~\cite{rusu2016progressive}. Lateral connections are performed to leverage knowledge from the older tasks to the current task. A drawback of these approaches is that their size grows rapidly incurring higher computational and storage costs.

PathNet is a deep learning architecture that uses agents embeddings in the network to look for nodes having features that can be reused by the current task~\cite{fernando2017pathnet}. The others that are specific to the previous tasks are fixed. In the learn to grow method, the structure and parameter optimization processes are performed after learning each task~\cite{li2019learn}. Various search criteria are exhausted to optimize the layers, nodes-per-layer and their weights. Regularize, Expand, and Compress mechanism applies a regularization technique and  expands the model architecture using an automatic neural architecture search engine~\cite{zhang2020regularize}. It applies a smart compression of the expanded model after a new task is learned to improve the model efficiency. Dynamically expanding network dynamically selects and duplicates the nodes that need to be retrained~\cite{yoon2017lifelong}. In the first place, it searches the nodes that are relevant to the current task and retrain them. If the model is not able to learn effectively with the previous step, its architecture expands by adding $K$ nodes to learn the features of the current task. The model weights that show a drastic deviation from the previous state are duplicated and the original ones are frozen before retraining the model again. \\

The plasticity-stability dilemma is an open challenge in memory-less continual deep neural networks. Increasing plasticity may reduce the model's stability and vice versa. The existing approaches make use of magnitude distance measures from the Minkowski measures  that do not provide better separation of nodes having large-sparse vectors~\cite{yoon2017lifelong}. These measures are too sensitive to feature scales, dimensionality of the data and their sparsity. Despite the optimal thresholds used, these measures present all points uniformly distant. 
For example, Let ${W}_1$ be the weight of a given node in the task $t_1$ and ${W}_2$ be the expected updated value of the corresponding nodes to accommodate task $t_2$. 
If ${W}_1 = [10, 10, 10, 10, 10]$ and ${W}_2 =[13, 13, 13, 13, 13]$, then the $dist({W}_1, {W}_2) = \|({W}_1 - {W}_2)\|_2^2 = (10 - 13)^2 + (10 - 13)^2 + (10 - 13)^2 + (10 - 13)^2 + (10 - 13)^2 = 45$. It is a high value but in actual the weights have increased in the same direction indicating a similar trend.

\section{Proposed Approach} \label{proposed}
We propose an approach that is more effective against semantic drift or catastrophic forgetting by better balancing of the plasticity and stability of the model. It is an improvement on the hybrid memory-less approach for continual deep learning that allows the model to regularize and expand as in~\cite{yoon2017lifelong}. It may also be referred to as the partial expansion method that ensures maximum utilization of the learnt weights and expanding dynamically for newer high-level features. Since the sequence of tasks are expected to be independent of each other, therefore, the model requires an automated mechanism to decide which nodes to freeze, regularize under constraint or expand through duplication. The model's performance heavily relies on objective function used to categorize nodes into three categories.  
There are mainly three steps to learn a newly introduced task. At first, a selective retraining block is performed to make use of similar sequential tasks~\cite{yoon2017lifelong}. Then by comparing the loss of the model with a given threshold, the model is dynamically expanded by K neurons to accommodate the new task's features~\cite{yoon2017lifelong, zhang2020regularize, li2019learn}. In the end, a neurons split or duplication operation is performed to prevent semantic drift while accommodating diverse features.

\begin{algorithm}
\caption{Proposed Approach}\label{alg:cap}
  \textbf{Input:} Task Sequence $T = T_1, T_2, ..., T_N$, Threshold $\sigma$ \\
 \textbf{Output:} $W_N$
\begin{algorithmic}[1]
\FOR{$T_i$ $\in$ $T$}
	\IF{$W_i$ is $\phi$}
    	\STATE  Train the model as first task, $W_1$ using Eq.~(\ref{eq:t1}) 
	\ELSE
		\STATE Load the trained model, $W_{i-1}$	
		\STATE Generate $W_i$ by retraining for $T_i$
		\FOR{all hidden nodes $j$}
             \STATE $\rho_{j}^{T_i} = \arccos(\frac{W_{j, T_i} . W_{j, T_{i-1}}}{||W_{j, T_i}||\times ||W_{j, T_{i-1}}||})\times (180/\pi)$
            \IF{$\rho_{j}^{T_i} > \sigma$}
                \STATE Duplicate the node j as $j^\prime$ and train for $W_i$
	        \ENDIF
        \ENDFOR
    \ENDIF
\ENDFOR
\end{algorithmic}
\end{algorithm}

Working of the proposed approach is represented in Algorithm~\ref{alg:cap}. It takes a sequence of $N$ tasks i.e., from $T_1$ till $T_N$ along with the threshold $\sigma$ for separating nodes . The output of the model is the weight matrix $W_N$ that supports all $N$ tasks. The lines 1~to~14 represent the processing of tasks in a sequence and may continue after $N$ as well. Lines 2 checks if this is the first task that the model is performing. In that case, the model is trained from scratch with the default architecture on line~3, using equation~\cite{yoon2017lifelong};

\begin{equation}\label{eq:t1}
\underset{W_{T_i}} {\text{minimize}} \mathcal{L}(W_1, T_1) + \mu \|W_1\|_1
\end{equation}

\noindent where $W_1$ are the network parameters after training on task $T_1$, while $\mu$ is the regularization parameter. In lines 4~to~13, the model updates its architecture for each task from $T_2$ till $T_N$. It first loads the parameter weights till the last task i.e., $W_{i-1}$ on line~5. At this stage the model is not trained for the current task $T_i$, however, the parameter weights $W_i$ is generated using a small fraction of the dataset for task $T_i$ using the same architecture, in line 6. It helps to evaluate the relevance between the current task and the previously performed tasks along with underlying semantic drift with respect to each node. The semantic drift is evaluated on lines 7~to~12 as the difference between $W_i$ and $W_{i-1}$ using the following equation;

\begin{equation}\label{eq:cosd}
\begin{split}
angular-dis(W_i, W_{i-1})  =  \\
  \arccos(\frac{\sum_{j=1}^J W_{1,j} \times W_{i-1,j}}{\sum_{j=1}^J W_{1,j} \times \sum_{j=1}^J W_{i-1,j}})\times \frac{180}{\pi}
\end{split}
\end{equation}

\noindent as angular distance between all the corresponding nodes from $j=1$ to $J$. Line~8, provide $\rho^{T_i}_j$ as the semantic drift in the node $j$ for the task $T_i$. In other words, the value of node $j$ will deviate by a margin of $\rho^{T_1}_j$ from its previous value if the new task is to be accommodated within the same architecture. In lines 9~to~11, this value is compared with the drift threshold $\sigma$. A value above the threshold indicates that changing the parameters of node $j$ for the current task $T_i$ would result in degrading the model's performance on the previously learnt tasks. Therefore, such nodes are duplicated in line~10 as $j^{\prime}$. It allows the model's architecture to expand and fit in the specific features of all learned tasks. The nodes with almost no semantic drift can be restrained from update to save computational cost. The updated architecture is fully trained for the task $T_i$ to get $W_i$ as;

\begin{equation}
\underset{W_i}{\text{minimize}} \mathcal{L}(W_i;T_i)+\lambda \|W_i-W_{i-1}\|_2^2
\end{equation}



\noindent The network weight matrix $W_i$ is obtained after modifying the model architecture and fully training the dataset for the task $T_i$. The weight matrix $W_{i-1}$ represents the parameter weights for the tasks $T_1$ till $T_{i-1}$. The $\lambda$ represents the penalization parameter that controls the distance between $W_{i}$ and $W_{i-1}$. To prevent catastrophic forgetting, a higher value for $\lambda$ enforce $W_{i}$ to be close in value to $W_{i-1}$. But if the sequential tasks are largely independent of each other, by using this regularization method, the model settles for a sub-optimal results. Therefore, the duplicate of such nodes are created and added adjacent to the corresponding nodes allowing better generalization across all tasks. Moreover, in the high dimensional feature space of deep neural networks, the angular distance provides better separation of nodes as compared to the magnitude based distance measures. It is robust against feature scales, sparsity and high dimensionality. It resulted in better accuracy for the proposed approach as compared to the existing approaches.


\begin{figure}
     \centering
     \begin{subfigure}[b]{0.3\textwidth}
         \centering
         \includegraphics[width=\textwidth]{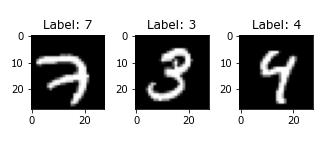}
         \caption{Original MNIST dataset samples}
         \label{fig:y equals x}
     \end{subfigure}
     \hfill
     \begin{subfigure}[b]{0.3\textwidth}
         \centering
         \includegraphics[width=\textwidth]{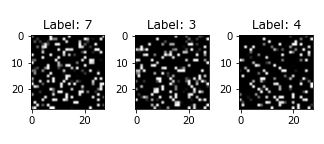}
         \caption{Permuted-MNIST dataset samples}
         \label{fig:three sin x}
     \end{subfigure}
     \hfill
     \begin{subfigure}[b]{0.3\textwidth}
         \centering
         \includegraphics[width=\textwidth]{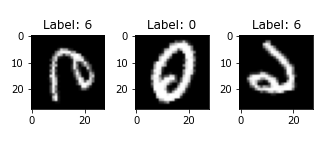}
         \caption{Rotated-MNIST dataset samples}
         \label{fig:five over x}
     \end{subfigure}
        \caption{Sample images from (a) Original MNIST (b) Permuted-MNIST and (c) Rotated-MNIST}
        \label{fig:datasets}
\end{figure}

\section{Experimental Setup and Results} \label{experiments}
The proposed approach is compared with state-of-the art continual deep learning models. All the models have a default architecture of two hidden layers having 312 and 128 units respectively. ReLu is being used as the activation function on both layers. The models are trained with an Adam optimiser and learning rate, batch-size and iterations are set to 0.001, 256 and 4300, respectively. The accuracy reported is the average accuracy across all the tasks, as in literature. 5-fold cross validation is used for reporting the performance measures. The other parameters of the existing models are set to their default values in the original work. 

In our experiments, we evaluated the performance of our proposed approach on the three datasets i.e., datasets permuted-MNIST, Permuted-Fashion-MNIST and Rotated-MNIST. They are widely used standard in continual deep learning. The permuted-MNIST and Permuted-Fashion-MNIST datasets are used as a sequence of 10 multi-class classification tasks, where in each task, the data has a different permutation. The Rotated-MNIST dataset is used as a sequence of 10 one-vs-rest binary classification tasks. The Permuted-MNIST dataset has 70K handwritten digital images or instances of 0 to 9 digits~\cite{MNIST}. Permuted-Fashion-MNIST also has 70K images of 10 different clothes from Zalando's article. Both of these datasets are generated by applying random permutations on the original MNIST and Fashion-MNIST datasets in order to make the detection task more challenging. Both datasets have 55K instances for training the model, 5K for validation and 10K for testing. Rotated-MNIST is generated by rotating the handwritten digits of the original MNIST to random angles between 0 and 360 degrees. Samples instances of the datasets are presented in \textbf{Fig.~\ref{fig:datasets}}.\\

\begin{figure}
    \centering
\begin{tikzpicture}
\begin{axis}[ybar, title={Average Accuracy after 10 tasks}, symbolic x coords={DEN, RWC, REC, EWC, Proposed},
  legend pos = north west, axis y line=none, axis x line=bottom, nodes near coords, enlarge x limits=0.2] 
\addplot+ coordinates {(DEN, 94.90) (RWC, 93.80) (REC, 95.70) (EWC, 84.40) (Proposed, 96.49)}; 
\end{axis} 
\end{tikzpicture}
\caption{A comparison of the proposed approach with latest models as average accuracy across 10 tasks on permuted-MNIST dataset}
\label{fig:comparison_all}
\end{figure}
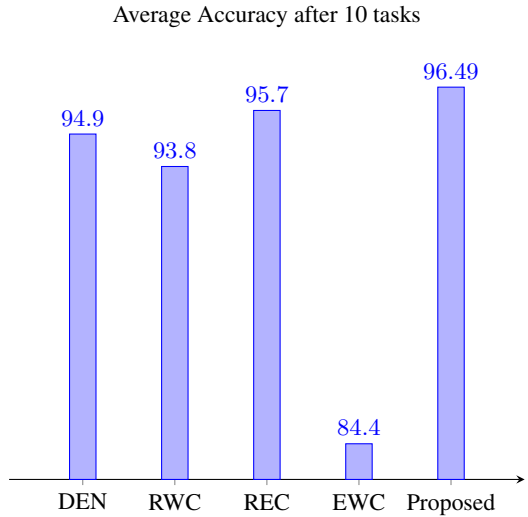

We have compared our proposed approach with the latest continual deep learning models on the permuted-MNIST dataset, as shown in \textbf{Fig.~\ref{fig:comparison_all}}. The proposed approach has outperformed all the existing models i.e., dynamically expendable networks (DEN), elastic weight consolidation (EWC), regularized weight consolidation (RWC) and regularize, expand and compress (REC)~\cite{zhang2020regularize}. 
Among these models, RWC~\cite{liu2018rotate} and EWC~\cite{kirkpatrick2017overcoming} are deep learning methods based on regularization to prevent semantic drift. While, DEN~\cite{yoon2017lifelong} is a partial expansion approach and REC~\cite{zhang2020regularize} is Regularize, Expand and Compress, a non-expansive continual learning method. When a single network was trained with the same setting for all the tasks, it resulted in an accuracy of 17.4\%. The proposed approach has improved accuracy by 0.79\% and 1.59\% as compared to second and third best models, respectively. The proposed approach has better objective function as angular distance, for separating nodes and adopts well to the other datasets as well.

\begin{table}
\caption{Performance of DEN and our proposed model with best thresholds ($\lambda$) and accuracy for the datasets Permuted-MNIST, Fashion-MNIST and Rotated-MNIST}
\label{table:intel1}
\begin{tabularx}{0.45\textwidth} { 
  | >{\raggedright\arraybackslash}X 
  | >{\centering\arraybackslash}X
  | >{\centering\arraybackslash}X
  | >{\raggedleft\arraybackslash}X | }
 \hline
 \textbf{Model} & \textbf{Dataset} & \textbf{$\lambda$} & \textbf{Accuracy} \\
 \hline
DEN & P-MNIST & 1 & 95.19\\
Proposed & & 30 & \textbf{96.49}\\
\hline
DEN & F-MNIST & 0.1 & 84.58\\
Proposed & & 30 & \textbf{86.01}\\
\hline
DEN & R-MNIST & 0.1 & 96.49\\
Proposed & & 20 & \textbf{97.66}\\
\hline
\end{tabularx}
\end{table}

DEN and the proposed approach belong to the category of partially expanding models that duplicate some nodes while partially regularize others. These models are well suited for tasks with higher diversity. The other approaches that do not expand, settle for sub-optimal solutions. In \textbf{Table~\ref{table:intel1}}, we report the competitive results of DEN and our proposed approach on the three datasets using the previously detailed setup. The hyper-parameter $\lambda$ is evaluated for different values, where only the ones resulting in highest accuracy are reported. This threshold is responsible for the stability-plasticity balance in dynamically expanding networks. The $\lambda$ values presented are scalars and angles for DEN and our proposed approach, respectively. Our proposed approach has outperformed DEN on all the three datasets.
For the Permuted MNIST dataset, the proposed approach with a hyper-parameter $\lambda$ of 30$^{\circ}$ yields the best performance. On the other hand, The best performance of the DEN is obtained with a hyper-parameter $\lambda$ of 1. Our proposed approach yields on average 1.3\% improvement in accuracy as compared to the DEN approach. For the Permuted Fashion MNIST dataset, our approach with a hyper-parameter $\lambda$ of 30$^{\circ}$ has the best performance over the rest and yields on average 1.43\% higher accuracy in comparison to the DEN approach. For the Rotated MNIST dataset, our approach with a hyper-parameter $\lambda$ of 20$^{\circ}$ yields the best performance again and gives on average 1.17\% accuracy improvement compared to the DEN approach. Overall, the proposed framework achieves around 1.3\% improvement in accuracy as compared to the prior method. \\



\section{Conclusion} \label{conclusion}
Continual deep learning models have many real-world applications. It allows pre-trained models to be fine-tuned for a number of tasks without the model forgetting about the previously learned tasks. The plasticity-stability dilemma is addressed differently in various existing models in order to balance retention of the previously acquired information and learning new information. The approaches that use dynamic expansion use a threshold to decide which nodes to freeze, partially regularize or duplicate based on the semantic drift for the corresponding task. Our proposed approach use angular distance to measure semantic drift as compared to scalar distance which has resulted in a higher balance between the stability and plasticity of the model. The proposed approach has outperformed the existing approaches across the standard continual deep learning datasets.  

\bibliographystyle{IEEEbib}
\bibliography{strings}

\end{document}